\documentclass[lettersize,journal]{IEEEtran}
\usepackage{amsmath,amsfonts}
\usepackage{algorithmic}
\usepackage{algorithm}
\usepackage{array}
\usepackage{textcomp}
\usepackage{stfloats}
\usepackage{url}
\usepackage{verbatim}
\usepackage{graphicx}
\usepackage{epstopdf}
\usepackage{cite}
\usepackage{booktabs}
\usepackage{adjustbox}
\usepackage{multirow}
\usepackage{hyperref}
\usepackage{mathtools}
\usepackage{adjustbox}
\usepackage{multirow}
\usepackage{enumitem}
\usepackage{multirow}
\usepackage{diagbox}
\usepackage{bm}
\usepackage{amsmath}  % 可以确保数学功能的可用
\usepackage{amssymb}  % 导入符号包
\usepackage{xcolor}
\usepackage{url}
\usepackage{pifont}
\begin{document}

\title{A Lightweight and Effective Image Tampering Localization Network with Vision Mamba}

\author{Kun Guo, Gang Cao, \IEEEmembership{Member, IEEE}, Zijie Lou, Xianglin Huang and Jiaoyun Liu  
\vspace{-0.5cm}
\thanks{
%Manuscript received xxxxxxxxxxx; revised xxxxxxxxxxx; accepted xxxxxxxxxxx. Date of publication xxxxxxxxxxx; date of current version xxxxxxxxxxx. This work was supported in part by the National Natural Science Foundation of China under Grant xxxxxxxx and in part by the Fundamental Research Funds for the Central Universities under Grant xxxxxxxx. The associate editor coordinating the review of this manuscript and approving it for publication was xxxxxxxxxxxxxxxxxx. (\textit{Corresponding author: Gang Cao.})
 
Kun Guo, Gang Cao, Zijie Lou and Xianglin Huang are with the School of Computer and Cyber Sciences, Communication University of China, Beijing 100024, China, and also with the State Key Laboratory of Media Convergence and Communication, Communication University of China, Beijing 100024, China (e-mail: \{kunguo, gangcao, louzijie2022,huangxl\}@cuc.edu.cn).

Jiaoyun Liu is with the School of Information Engineering, Changsha Medical University, Changsha 410219, China (e-mail: 397507500@qq.com).

%Code is available at \href{https://github.com/multimediaFor/ForMa}{https://github.com/multimediaFor/ForMa.}

%Digital Object Identifier xxxxxxxxxxxxxxxxx
}
}

\markboth{Journal of \LaTeX\ Class Files, Vol. 14, No. 8, August 2015}
{Shell \MakeLowercase{\textit{et al.}}: Bare Demo of IEEEtran.cls for IEEE Journals}
\maketitle

\begin{abstract}

Current image tampering localization methods primarily rely on Convolutional Neural Networks (CNNs) and Transformers. While CNNs suffer from limited local receptive fields, Transformers offer global context modeling at the expense of quadratic computational complexity. Recently, the state space model Mamba has emerged as a competitive alternative, enabling linear-complexity global dependency modeling. Inspired by it, we propose a lightweight and effective FORensic network based on vision MAmba (ForMa) for blind image tampering localization. Firstly, ForMa captures multi-scale global features that achieves efficient global dependency modeling through linear complexity. Then the pixel-wise localization map is generated by a lightweight decoder, which employs a parameter-free pixel shuffle layer for upsampling. Additionally, a noise-assisted decoding strategy is proposed to integrate complementary manipulation traces from tampered images, boosting decoder sensitivity to forgery cues. Experimental results on 10 standard datasets demonstrate that ForMa achieves state-of-the-art generalization ability and robustness, while maintaining the lowest computational complexity. Code is available at \href{https://github.com/multimediaFor/ForMa}{https://github.com/multimediaFor/ForMa.}

\end{abstract}


\IEEEpeerreviewmaketitle

\section{Introduction}

With the development of image editing tools and technologies, users can easily manipulate images without requiring extensive professional knowledge. If used maliciously, such technologies pose a threat to social stability and harmony\cite{spl2024lightweight}. It also presents challenges to the problem of image forgery localization task \cite{guo2024effective,zhu2024ConvNextFF}, which is aimed at discovering the specific altered regions within a forged image.

\begin{figure}[!t]
    \centering
    \includegraphics[width=0.85\linewidth]{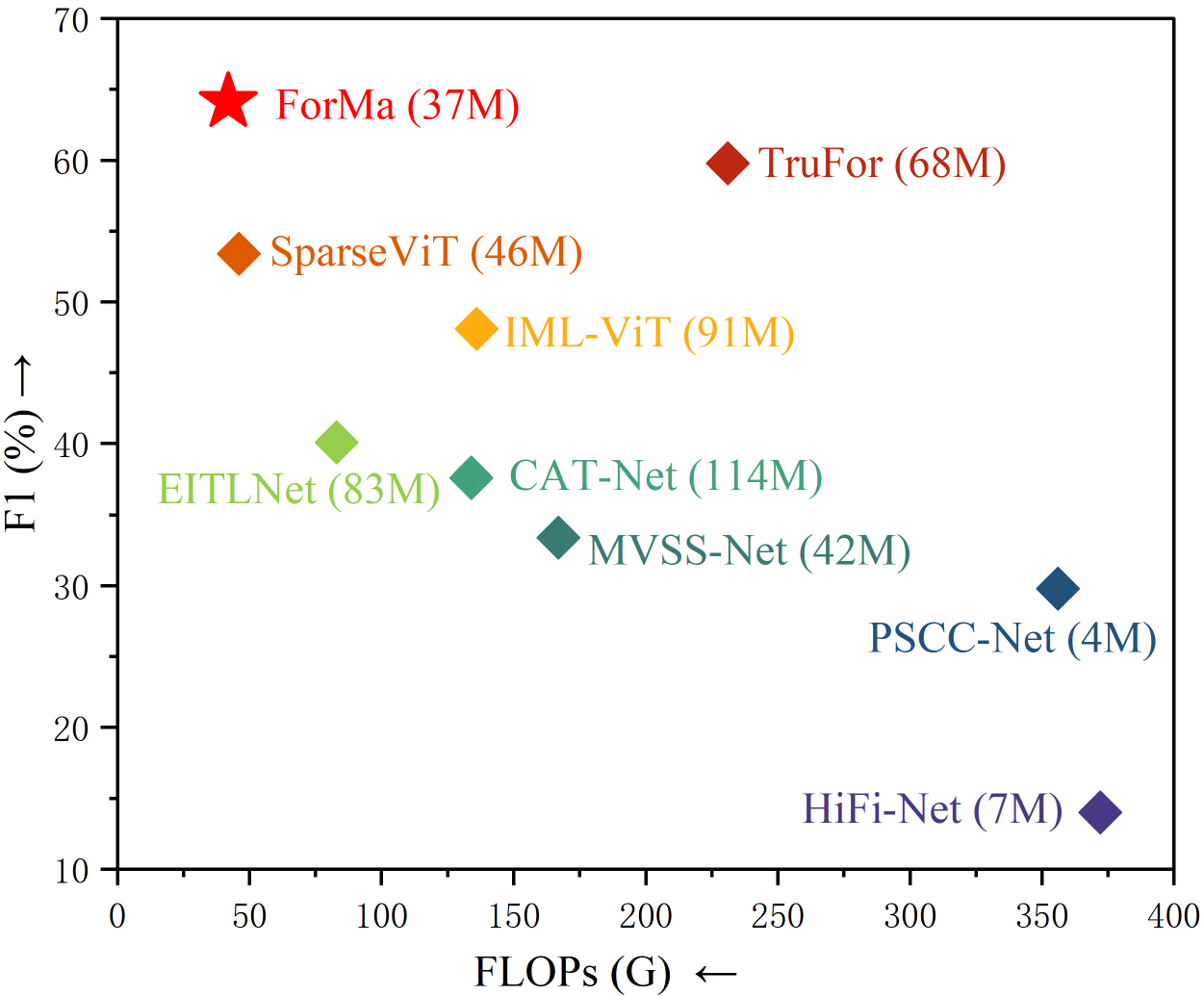}
    \caption{Comparison of average F1 across 10 standard benchmark datasets and FLOPs (calculated in 512 × 512 input size) with model parameters. Our method achieves the best F1 (64.1\%) with the lowest FLOPs (42G) and parameters (37M).}
    \label{fig:bubble}
\vspace{-0.5cm}
\end{figure}

In recent years, various deep learning-based  methods have been proposed to achieve image forgery localization. Early methods are primarily based on CNN architectures, such as MVSS-Net \cite{dong2022mvss}, CAT-Net \cite{kwon2022learning}, PSCC-Net \cite{liu2022pscc}, and HiFi-Net \cite{hifi_net_xiaoguo}. Later, Transformer-based architectures with attention mechanism emerge as promising approaches, such as EITLNet \cite{guo2024effective}, TruFor \cite{guillaro2023trufor}, and IML-ViT \cite{ma2023imlvit}. As shown in Fig. \ref{fig:bubble}, CNN-based methods generally exhibit low localization accuracy and poor generalization, with F1 ranging from 14.4\% to 37.6\%. Due to the ability of attention mechanism in capturing global information, the Transformer-based methods achieve F1 within [40.1\%, 59.8\%], which are significantly higher than those of the CNN methods. However, such performance improvement comes at the cost of increased parameters and computational complexity. For instance, the best Transformer-based method, i.e., TruFor improves the F1 from 37.6\% to 59.8\% compared to the best CNN-based method, i.e., CAT-Net, but such FLOPs also increase from 134G to 231G. The computation and storage of these parameters place higher demands on hardware, making it difficult to deploy such forensic methods on standard hardware devices. 
% Although recent methods, such as SparseViT \cite{su2025sparse} and LoMa\cite{lou2024loma}, have addressed this issue by employing sparse vision transformers or SSM structures to reduce the number of parameters during training, they sacrifice some localization accuracy in the process.

In this letter, we propose ForMa, a lightweight and effective image tampering localization network. Benefiting from the foundational research on classical State Space Sequence models (SSMs) \cite{kalman1960new}, modern SSM architectures such as Mamba \cite{gu2023mamba} not only establish long-range dependencies with strong feature representation capabilities but also exhibit linear complexity with respect to input size. We utilize VMamba \cite{liu2024vmamba} as ForMa's backbone to extract multi-scale features. ForMa incorporates a lightweight decoder merely composed of linear layers, which employs pixel shuffle operations to further decrease computational costs. Additionally, a noise-assisted decoding strategy is integrated to extract auxiliary forensic features, enhancing our method's ability to accurately locate tampered regions. Compared to existing state-of-the-art CNN- and Transformer-based approaches, ForMa achieves superior localization accuracy, while significantly reduces both parameters and computational complexity.

The rest of this letter is organized as follows. The proposed ForMa scheme is described in Section II, followed by extensive experiments and discussions in Section III. We draw the conclusions in Section IV.

\section{Proposed Method}
\begin{figure*}[!t]
    \centering
    \includegraphics[width=\textwidth]{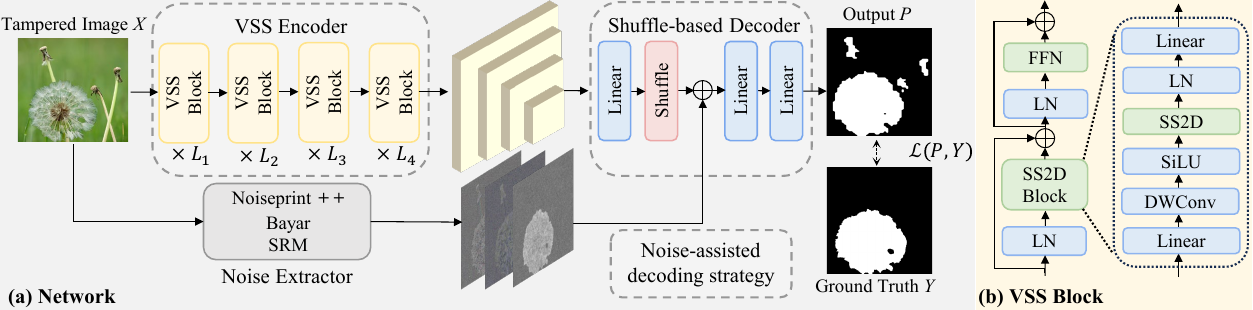}
    \caption{(a) Architecture of proposed ForMa. $L_i$=\{2, 2, 9, 2\}. Linear, Conv, and Shuffle refers to the linear, convolution and pixel shuffle layers, respectively. $\oplus$ represents element-wise addition. (b) Structure of the VSS Block. It includes a depthwise convolutional layar (DWConv), SiLU activation function, SS2D module, and linear normalization (LN). The VSS Block, Shuffle-based Decoder, and Noise Extractor are learnable.}
    \label{fig:framework}
\end{figure*}

\begin{figure}[!t]
    \centering
    \includegraphics[width=\linewidth]{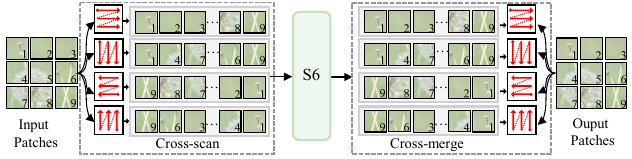}
    \caption{Illustration of the SS2D module. The S6 used is from Mamba\cite{gu2023mamba}.}
    \label{fig:vss}
\end{figure}
\label{sec:method}
As depicted in Fig. \ref{fig:framework}, the proposed ForMa architecture presents three key innovations: (1) a visual state space (VSS) encoder replacing conventional CNN/Transformer backbones, (2) a novel lightweight decoder with pixel shuffle-based upsampling and (3) a noise-assisted decoding strategy. 

Given an input RGB image $X \in \mathbb{R}^{H \times W \times 3}$, ForMa processes it through VSS blocks. This generates hierarchical feature maps ${F_i}$ $(i=1,2,3,4)$ at progressively reduced spatial resolutions $\frac{H}{2^{i+1}} \times \frac{W}{2^{i+1}}$ with channel dimensions $C_i$, effectively capturing multi-scale contextual information through state space modeling. Unlike computationally heavy decoder designs in prior works\cite{guillaro2023trufor, guo2024effective}, we implement an efficient upsampling mechanism via pixel shuffle. This operation progressively recover spatial details through channel reorganization, achieving a favorable balance between parameter efficiency and detail preservation. Meanwhile,  a noise feature extractor integrates the noise residual features extracted from Noiseprint++\cite{guillaro2023trufor}, SRM\cite{wu2019mantra}, and Bayar Convolution\cite{dong2022mvss}, which boosts decoder sensitivity to manipulation cues.

\subsection{VSS Encoder with SS2D Module}
\label{subsec:S6}
Traditional State Space Models\cite{gu2021efficiently-s4} map 1D input sequences $x(t)$ to outputs $y(t)$ via a latent state $h(t) \in \mathbb{R}^N$, governed by static parameters ($\mathbf{\Delta}$,$\mathbf{A},\, \mathbf{B},\,\mathbf{C}$). Such fixed parameters limit the adaptability to dynamically varying sequences. Mamba\cite{gu2023mamba} addressed this with the Selective State Space Model (S6), introducing input-dependent dynamics. The S6 system can be expressed as:
\begin{equation}
\label{eq:s6}
\begin{aligned}
h(t) &= \overline{\mathbf{A}}h(t-1) + \overline{\mathbf{B}}x(t), \quad y(t) = \mathbf{C}h(t)
\end{aligned}
\end{equation}
where $\overline{\mathbf{A}} \in \mathbb{R}^{N\times N} = f_\mathbf{A}(\mathbf{\Delta}, \mathbf{A})$, $\overline{\mathbf{B}} \in \mathbb{R}^{N \times 1} = f_\mathbf{B}(\mathbf{\Delta}, \mathbf{A}, \mathbf{B})$, and $\mathbb{\mathbf{C}} \in \mathbb{R}^{1 \times N}$ are parameter matrices.  $f_\mathbf{A}$  and $f_\mathbf{B}$  are the discretization functions for transforming $\mathbf{A}$ and $\mathbf{B}$ into discrete $\overline{\mathbf{A}}$ and $\overline{\mathbf{B}}$, respectively. To apply S6 to the image tampering localization task, the images must be serialized. As shown in Fig. \ref{fig:vss}, the 2D Selective Scan (SS2D)\cite{liu2024vmamba} integrates spatial context by first unfolding input patches into sequences along four directions via Cross-scan, processing each with parallel S6 systems, then merging outputs through Cross-merge to restore spatial dimensions. Our proposed ForMa captures multi-scale contextual information through above state space modeling.

\subsection{ Noise-assisted Decoding Strategy }
Previous image forensics approaches typically combine noise features with RGB inputs through early fusion for joint encoder processing \cite{wu2019mantra, guo2024effective, guillaro2023trufor}. Our methodology diverges by strategically incorporating noise features into the decoder stage as auxiliary localization cues, rather than employing early fusion at the encoder input. Such noise-assisted decoding strategy enhances tampering localization precision while maintaining encoder efficiency. The effectiveness of such design is validated via ablation studies in Section \ref{ablation study}.
Recognizing that single-source noise inadequately adapts to diverse manipulation types, we integrate forensic features extracted from SRM\cite{fridrich2012rich}, Noiseprint++\cite{guillaro2023trufor}, and Bayar convolution\cite{dong2022mvss} as complementary foresic artifacts. Specifically, the three forensic features are concatenated along the channel dimension and processed through a convolutional block to generate fused features $F_{mod}$ with spatial resolution $\frac{H}{4} \times \frac{W}{4} \times {C_{mod}}$.

% \begin{equation}
%     \begin{aligned}
%     {{F}_m = {Conv}(R_{mod})}, mod\in \{noiseprint,srm,bayar\}
%     \end{aligned}
%     \label{eq:extracor1}
% \end{equation}
% where $R_{noisprint}$,\,$R_{srm}$,\,$R_{bayar}$ represent the noise features of the RGB image $x$ extracted from the NoisePrint++, SRM and Bayar filters respectively.

\subsection{Lightweight Shuffle-based Decoder}
ForMa's architecture integrates a shuffle-based decoder that optimizes computational complexity through parameter-free upsampling.
Firstly, the multi-scale feature maps obtained from the encoder are individually processed through an linear layer with an expansion factor $r_i$, which denotes the feature map magnification ratio.  We define $r_i$ as $[1,2,4,8]$, such can be formulated as:
\begin{equation}
    \begin{aligned}
    {\hat{F}_i = \text{$Linear$}(C_i, C \times r_i^2 )(F_i)},\forall i
    \end{aligned}
    \label{eq:decoder1}
\end{equation}
where  $\text{$Linear$}({C_{\text{in}}}, {C_{\text{out}}})(\cdot)$  refers to a linear layer taking a $C_{\text{in}}$-dimensional tensor as input and generating a $C_{\text{out}}$-dimensional tensor as output. $C=96$ specifies the default embedding tensor dimension.  Next, we introduce the pixel shuffle as the upsampling layer. Unlike the bilinear interpolation common used in previous methods \cite{guo2024effective, guillaro2023trufor}, pixel shuffle is a parameter-free and effective method that converts channel dimensions into spatial dimensions, thereby further reducing the computational complexity of the localization network. By applying pixel shuffle, we obtain new feature maps \( {\hat{F}}_i \) with spatial dimensions of \( \frac{H}{4} \times \frac{W}{4} \). Subsequently, \( {\hat{F}}_i \) is concatenated with the features \( F_{mod} \) extracted by the noise extractor, and passed through a linear layer for fusion. Finally, a new linear layer is used to perform pixel-level prediction on the fused features, generating the predicted output $P$. This can be formulated as:
\begin{equation}
    \begin{aligned}
    &{{\hat{F}_i} = \text{$Shuffle$}({r_i})({\hat{F}}_i)},\forall i \\
    &{{F} = \text{$Linear$}(4C+C_{mod}, C)(\text{$Concat$}({{\hat{F}_i}},{F}_{mod}})),\forall i\\
    &{{P} = \text{$Linear$}(C, 2)({F})}
    \end{aligned}
    \label{eq:decoder2}
\end{equation}
where  $\text{$Shuffle$}( {r_i})(\hat{F}_i)$  refers to employing a pixel shuffle layer to scale the height and width of $\hat{F_i}$ by a factor of $r_i$, respectively. $\text{$Concat$}{({{\hat{F}_{i}}},{F}_{mod}})$ denotes to merges features channel-wise.

% \subsection{Lightweight Shuffle-based Decoder} 
%  $C=96$ specifies the default embedding dimension. The subsequent upsampling phase replaces conventional bilinear interpolation \cite{guo2024effective, guillaro2023trufor} with pixel shuffle operation. Such parameter-free method reshapes channel dimensions into spatial dimensions, producing new feature maps \( {\hat{F}}_i \in \mathbb{R}^{\frac{H}{4} \times \frac{W}{4} \times C}\) while reducing computational overhead. Subsequently, \( {\hat{F}}_i \) is concatenated with the features \( F_{mod} \) extracted by the noise extractor, and passed through a linear layer for fusion. Finally, a new linear layer is used to perform pixel-level prediction on the fused features, generating the predicted output $P$. This can be formulated as:

\begin{table*}
\caption{Image forgery localization performance F1[\%] and IoU[\%]. The corresponding number of images is annotated for each test set. The best results of per test set are highlighted in \textcolor{red}{\textbf{red}}.}
\begin{adjustbox}{width=\textwidth}
\begin{tabular}{r|c|cccccccccccccccccccc|cc}
\toprule[1.5pt]
\multirow{2}{*}{Method} &\multirow{2}{*}{Architecture} & \multicolumn{2}{c}{Columbia (160)} & \multicolumn{2}{c}{DSO (100)} & \multicolumn{2}{c}{CASIAv1 (920)} & \multicolumn{2}{c}{NIST (564)} & \multicolumn{2}{c}{Coverage (100)}& \multicolumn{2}{c}{Korus (220)}& \multicolumn{2}{c}{Wild (201)} & \multicolumn{2}{c}{CoCoGlide (512)} & \multicolumn{2}{c}{MISD (227)}  & \multicolumn{2}{c|}{FF++ (1000)} & \multicolumn{2}{c}{\textbf{Average}} \\ \cmidrule{3-24}
                         & & F1  & IoU       & F1    & IoU      & F1   & IoU         & F1   & IoU       & F1   & IoU  & F1 & IoU     & F1    & IoU     & F1     & IoU      & F1     & IoU   & F1     & IoU  & F1     & IoU\\         
\midrule
PSCC \cite{liu2022pscc}  &CNN &61.5 &48.3 &41.1 &31.6 &46.3 &41.0 &18.7 &13.5 &44.4 &33.6  & 10.2 &5.8 &10.8 &8.1 &42.1 &33.2 &65.6 &52.4   &7.0  &4.2 &29.8 &23.9       \\
MVSS-Net \cite{dong2022mvss}  &CNN &68.4 &59.6 &27.1 &18.8 &45.1 &39.7 &29.4 &24.0 &44.5 &37.9  &9.5 &6.7 &29.5 &21.9 &35.6 &27.5 &65.9 &52.4   &16.5  &12.7 &33.4 &27.4       \\
HiFi-Net \cite{hifi_net_xiaoguo}  &CNN &83.3 &74.1 &11.3 &7.5 &9.7 &7.8 &12.8 &7.8 &10.5 &6.3  &8.7 &5.5 &3.8 &2.3 &21.1 &14.8 &24.4 &17.4   &7.1  &4.2 &14.4 &10.5       \\
CAT-Net \cite{kwon2022learning}    &CNN &79.3 &74.6 &47.9 &40.9 &71.0 &63.7 &30.2 &23.5 &28.9 &23.0  & 6.1 &4.2 &34.1 &28.9 &36.3 &28.8 &39.4 &31.3    &12.3  &9.5 &37.6 &32.0       \\
\midrule 
EITLNet \cite{guo2024effective}   &Transformer   &87.6 &84.2 &42.2 &33.0 &55.7 &52.0 &33.0 &26.7 &44.3 &35.3  &32.3&26.1&51.9 &43.0 &35.4 &28.8 &75.5 &63.8   &15.1  &10.7&40.1 &34.3   \\
IML-ViT \cite{ma2023imlvit}   &Transformer   &91.4 &88.9 &14.4 &9.0 &81.1 &74.9 &8.8 &6.6 &17.5 &11.9  &6.5&3.9&30.4 &24.0 &31.5 &24.5 &66.6 &53.5   &56.4  &47.6&48.1 &41.7   \\
SparseViT \cite{su2025sparse}  &Transformer &\textcolor{red}{\textbf{95.8}} &\textcolor{red}{\textbf{94.8}} &24.5 &21.2 &\textcolor{red}{\textbf{81.9}} &\textcolor{red}{\textbf{76.8}} &38.3 &32.3 &51.2 &46.9  &20.8 &16.5 &50.1 &44.5 &38.6 &32.4 &\textcolor{red}{\textbf{76.2}} &\textcolor{red}{\textbf{63.7}}  &42.3  &34.3 &53.4 &47.2       \\
TruFor \cite{guillaro2023trufor}   &Transformer   &79.8 &74.0 &\textcolor{red}{\textbf{91.0}} &\textcolor{red}{\textbf{86.5}} &69.6 &63.2 &\textcolor{red}{\textbf{47.2}} &\textcolor{red}{\textbf{39.6}} &52.3 &45.0  &\textcolor{red}{\textbf{37.7}}&\textcolor{red}{\textbf{29.9}}&\textcolor{red}{\textbf{61.2}} &\textcolor{red}{\textbf{51.9}} &35.9 &29.1 &60.0 &47.5    &69.2  &56.5&59.8 &51.1   \\
\midrule 
%LoMa                      &Mamba &88.9 &85.2 &40.7 &29.5 &{76.6}&69.9 &45.9 &38.5 &52.1 &43.0  & 27.5&21.5&54.2 &44.2 &43.2 &33.7 &68.2 &55.3   &71.9  &60.8&61.5 &52.7         \\
Ours                     &Mamba &94.9 &93.9 &38.7 &28.3 &72.9 &67.3 &45.4 &38.5 &\textcolor{red}{\textbf{58.7}} &\textcolor{red}{\textbf{50.9}}  & 30.4&23.5&57.3 &48.7 &\textcolor{red}{\textbf{45.3}} &\textcolor{red}{\textbf{36.2}} &70.3 &57.5   &\textcolor{red}{\textbf{81.9}}  &\textcolor{red}{\textbf{71.8}}&\textcolor{red}{\textbf{64.1}} &\textcolor{red}{\textbf{56.2}}        \\

\bottomrule[1.5pt]
\end{tabular}
\end{adjustbox}
\label{800k}
\end{table*}

\section{Experiment}
\label{experiment}
\subsection{Experimental Settings}

\noindent
 \textbf{Datasets.} Consistent with CAT-Net \cite{kwon2022learning}, TruFor \cite{guillaro2023trufor}, IML-ViT\cite{ma2023imlvit} and SparseViT\cite{su2025sparse}, our ForMa is trained on the CAT-Net dataset \cite{kwon2022learning}. Such a training dataset contains over 800\textit{k} forged images with diverse tampering types, such as splicing, copy-move and object removal. For measuring the generalization ability, ten cross-domain datasets are used in tests, where no overlap exists between the training and test datasets. The test datasets include CASIAv1 \cite{dong2013casia}, Columbia \cite{hsu2006columbia}, NIST \cite{guan2019mfc}, DSO \cite{de2013exposing}, Coverage \cite{wen2016coverage}, Korus\cite{korus2016evaluation}, Wild \cite{huh2018fighting}, MISD\cite{kadam2021multiple}, FF++\cite{rossler2019faceforensics++} and CoCoGlide\cite{guillaro2023trufor}.

\noindent
 \textbf{Implementation Details.} 
 The network is initialized by the weights VMamba-tiny\cite{liu2024vmamba} pretrained on ImageNet. The training is performed on an A100 GPU 40GB with the batch size 8, and all the images are resized to $512 \times 512$ pixels. The learning rate is adjusted from 1e-4 to 1e-8 by ReduceLROnPlateau decay strategy. AdamW is adopted as the optimizer with default momentum parameters $(\beta_1=0.9, \beta_2=0.999)$. The common data augmentations, including flipping, blurring, compression and noising are adopted.

\noindent
 \textbf{Evaluation Metrics.} As previous works  \cite{guillaro2023trufor, guo2024effective,dong2022mvss}. F1 and IoU are used as evaluation criteria with a default threshold 0.5. Considering the obvious imbalance of image sample numbers across different datasets, we follow the approach outlined in \cite{lou2024exploring} to calculate the average F1 and IoU. 
 
\noindent
 \textbf{Loss Function.} In line with \cite{guo2024effective}, we utilize a combined loss consisting of the DICE\cite{dong2022mvss} and Focal losses\cite{focalloss2017}.

\subsection{Comparison with State-of-the-Arts}

\begin{figure}[!t]
    \centering
    \includegraphics[width=\linewidth]{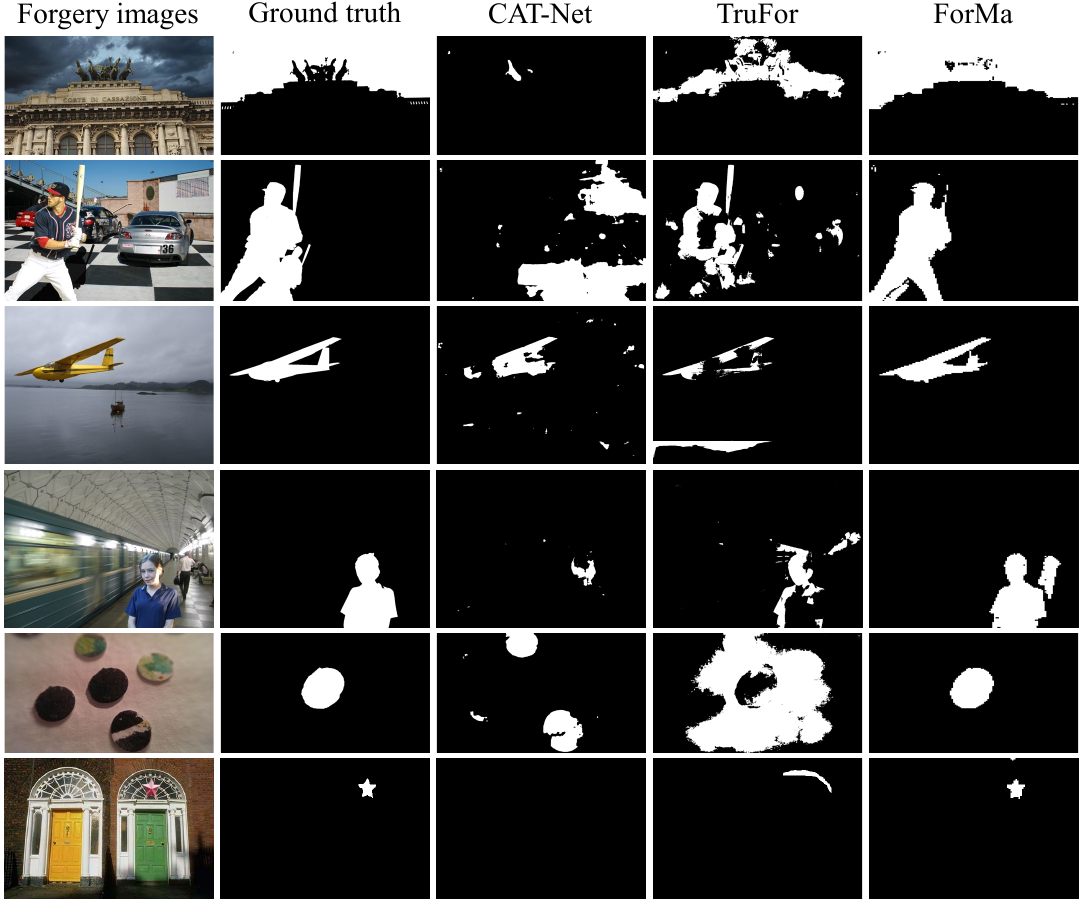}
    \caption{Results on example images from NIST, Korus and CASIAv1 datasets. From left to right: tampered images, ground truth, localization results from CAT-Net (the best CNN-based method), TruFor (the best Transformer-based method) and ForMa.}
    \label{fig:visual}
    \vspace{-0.3cm}
\end{figure}

Table \ref{800k} presents a quantitative comparison of 8 state-of-the-art image forgery localization methods.  Our method achieves the best average performance across 10 datasets in terms of F1 and IoU, reaching 64.1\% and 56.2\%, respectively.  ForMa surpasses  TruFor (the best Transformer-based method) by F1 +4.3\% and IoU +4.9\%, as well as CAT-Net (the best CNN-based method) by F1 +6.5\%, IoU +24.2\%.  Such outperformance demonstrates the superiority of our method. On the FF++ (DeepFake) and CoCoGlide (diffusion model generation) datasets, ForMa achieves F1 of 81.9\% and 45.3\% respectively, demonstrating its strong generalization ability. Fig. \ref{fig:visual} provides an intuitive insight into the tampering localization results. It can be seen that ForMa exhibits fewer false alarms and missing detection of the tampered regions.

In terms of computational complexity, our method achieves the lowest cost. As shown in Table \ref{flops}, ForMa attains 42G FLOPs for 512×512 images and 70G FLOPs for 1024×1024 images, both of which are the lowest among existing methods. Additionally, our method is maintaining a low level of 37M parameters. In summary, our proposed ForMa not only outperforms existing methods in terms of localization accuracy but also achieves a significant reduction in computational complexity. %Such sufficiently demonstrates the lightweight and effectiveness of ForMa.

% =============================================================================
\begin{table}[!t]
\caption{Complexity of Our method compared to SoTA models. The lowest computational complexity is highlighted in \textcolor{red}{\textbf{red}}.}
\tabcolsep=9 pt
\centering
\begin{adjustbox}{width=\linewidth}
\begin{tabular}{r|l|ccc}
\toprule
\multirow{2}{*}{Method} &\multirow{2}{*}{Year-Venue}  & \multirow{2}{*}{Params. (M)} & \multirow{2}{*}{\begin{tabular}[c]{@{}c@{}}512×512 \\ FLOPs (G)\end{tabular}} & \multirow{2}{*}{\begin{tabular}[c]{@{}c@{}}1024×1024\\ FLOPs (G)\end{tabular}} \\
                        &                              &                                                                               &                                                                                \\
\midrule
PSCC \cite{liu2022pscc} &2022-TCSVT & 3           & 356          & 649               \\
MVSS-Net \cite{dong2022mvss}  &2022-TPAMI& 147            & 167          & 683                \\
HiFi-Net \cite{hifi_net_xiaoguo}  &2023-CVPR                & 7            & 372          & 3342                \\
CAT-Net \cite{kwon2022learning}  &2022-IJCV                & 114            & 134          & 538                \\
\midrule
EITLNet \cite{guo2024effective} &2024-ICASSP & 52            & 83                        & 426                    \\
IML-ViT \cite{ma2023imlvit}  &2024-AAAI& 91            & 136                        & 445                    \\
SparseViT \cite{su2025sparse}  &2025-AAAI& 50            & 46                        & 185                    \\
TruFor \cite{guillaro2023trufor} &2023-CVPR                  & 69              & 231       & 1016               \\
\midrule
% LoMa &2025-arxiv                  & 37              & 64       & 258               \\
Ours       &2025             & 37            & \textcolor{red}{\textbf{42}}                      & \textcolor{red}{\textbf{170}}                    \\
\bottomrule
\end{tabular}
\end{adjustbox}
\label{flops}
\vspace{-0.2cm}
\end{table}
% ===========================================================================

\subsection{Ablation Studies}
\label{ablation study}
To assess the impact of key design components on localization accuracy, several ablation experiments are conducted. As shown in Table \ref{ablation}, the results show that combining Focal loss and DICE loss improves F1 on CoCoGlide by at least 9.3\%, and IoU by at least 7.4\%, respectively, confirming the advantage of using combined losses. The noise extractor improves performance across all datasets, with a notable 16.1\% increase in F1 and 17\% in IoU on FF++. Pixel shuffle increases F1 by 9.0\% and IoU by 10.5\% on FF++, and 3.5\% and 2.3\% on Korus. By placing the noise extractor in the decoding process rather than employing early fusion with RGB images for encoder feature extraction, our method achieves a 6.7\% F1 improvement on  Korus dataset and reduces FLOPs by 192 G. Overall, such results demonstrate the effectiveness of the individual components comprising the proposed ForMa.
\vspace{-0.4cm}
\begin{table}[!t]
\caption{Ablation analysis of our proposed scheme. Metric values are in percentage. The best results are highlighted in \textcolor{red}{\textbf{red}}.}
\tabcolsep=5 pt
\centering
\begin{adjustbox}{width=\linewidth}
\begin{tabular}{lcccccccccccccc}
\toprule
\multirow{2}{*}{Methods}  & \multicolumn{2}{c}{Korus} & \multicolumn{2}{c}{CoCoGlide} & \multicolumn{2}{c}{FF++}&\multicolumn{1}{c}{512$\times$512} \\
\cmidrule(r){2-3}  \cmidrule(r){4-5} \cmidrule(r){6-7} 
\multicolumn{1}{c}{}   & F1            & IoU          & F1         & IoU        & F1           & IoU      & FLOPs (G)          \\ \midrule
 w/o Focal Loss                 & 26.2         & 20.3        & 36.0      & 28.8      & 73.4        & 61.9      &42         \\
 w/o DICE Loss              &  {15.8}         &  {12.5}        & 20.6      & 16.4      & 48.8        & 37.5       &42        \\
 w/o Noise Extractor              &  {26.1}         &  {20.5}        & 43.3      & 34.9      & 65.8        & 54.8   &30          \\
 w/o Shuffle             &  {26.9}         &  {21.3}        & 40.7      & 32.5      & 72.9        & 61.3      &68       \\
 Noise fed into Encoder            &  {23.7}         &  {18.3}        & 41.0      & 32.6      & 76.4        & 64.6      &232       \\
ForMa          & \textcolor{red}{\textbf{30.4}}        & \textcolor{red}{\textbf{23.5}}        & \textcolor{red}{\textbf{45.3}}      & \textcolor{red}{\textbf{36.2}}      & \textcolor{red}{\textbf{81.9}}       & \textcolor{red}{\textbf{71.8}} &\textcolor{red}{\textbf{42}}\\
\bottomrule
\end{tabular}
\end{adjustbox}
\label{ablation}
\vspace{-0.5cm}
\end{table}

\subsection{Robustness Evaluation}
\begin{figure}[!t]
    \centering
    \includegraphics[width=\linewidth]{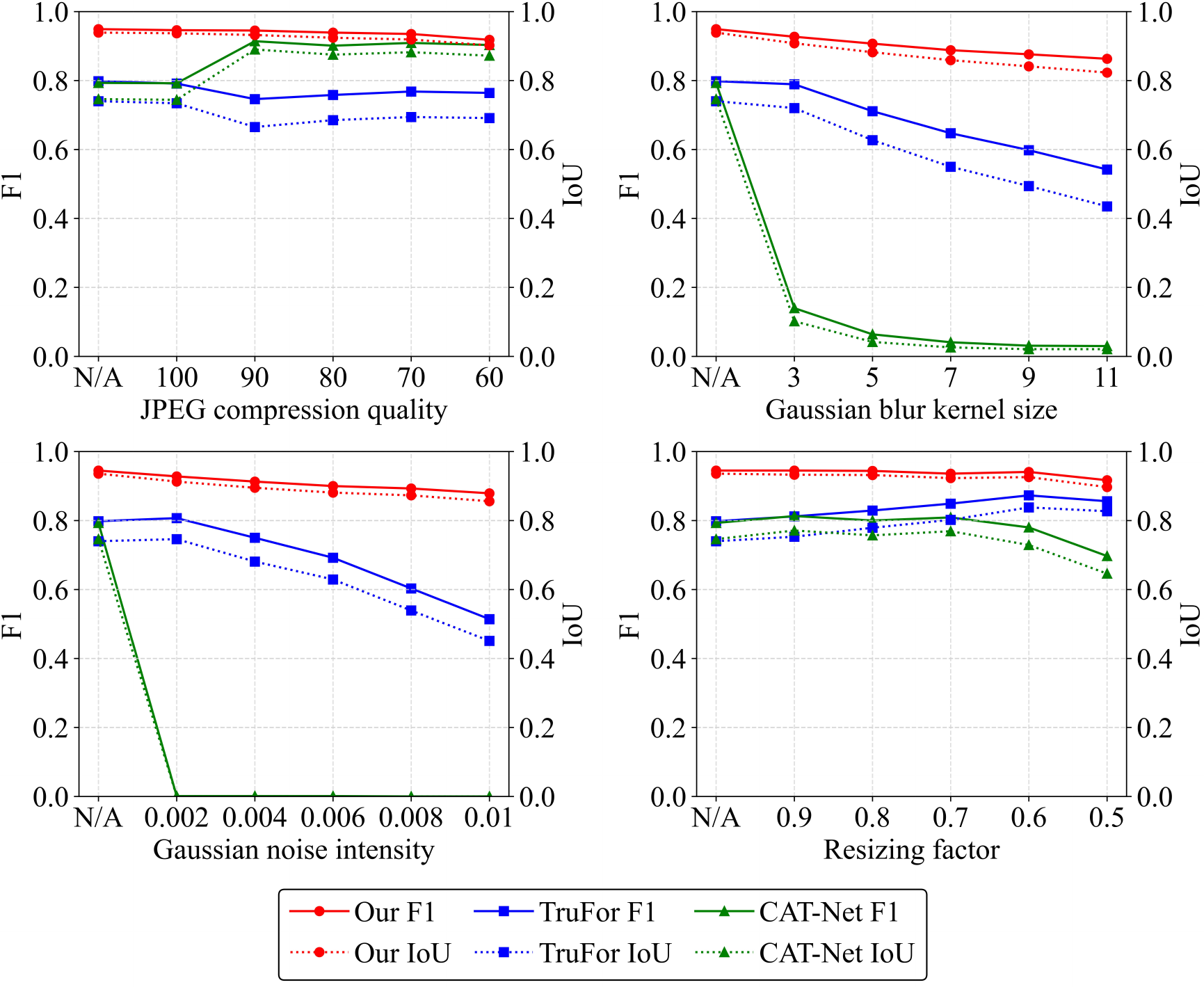}
    \caption{Robustness evaluation against different image post-processing techniques on Columbia dataset.}
    \label{robust}
    \vspace{-0.3cm}
\end{figure}

 We first assess the robustness of our model against post-processing effects from online social networks (OSNs) such as Facebook, Weibo, Wechat, and Whatsapp \cite{wu2022robust}. As shown in Table \ref{osn}, ForMa maintains a performance advantage even after severe post-processing. On the Facebook platform, our F1  achieves 70.3\%,  higher than 67.3\% F1 of  TruFor and 63.3\% F1 of CAT-Net. Moreover, our method demonstrates stability across various online platform post-processing scenarios. On the NIST dataset,  our method consistently outperforms TruFor across all platforms. For instance, we get F1 44.8\% on Weibo while TruFor gets 33.2\%. Following the prior works\cite{dong2022mvss, liu2022pscc}, the robustness against the post JPEG compression, Gaussian blur, Gaussian noise and resizing is also evaluated on Columbia dataset.  As shown in Fig. \ref{robust}, the ForMa always behaves the best. It verifies the high robustness of our scheme against such post manipulations.

\begin{table}[!t]
\caption{Robustness performance F1[\%] and IoU[\%] against online social networks (OSNs) post-processing, including Facebook (FB), Wechat (WC), Weibo (WB) and Whatsapp (WA). The best results are highlighted in \textcolor{red}{\textbf{red}}.}
\centering
\tabcolsep=5 pt
\begin{adjustbox}{width=\linewidth}
\begin{tabular}{r|c|cccccccc|cc}
\toprule
\multirow{2}{*}{Method} & \multirow{2}{*}{OSNs}     & \multicolumn{2}{c}{CASIAv1} & \multicolumn{2}{c}{Columbia} & \multicolumn{2}{c}{NIST} & \multicolumn{2}{c|}{DSO} & \multicolumn{2}{c}{Average}  \\ \cmidrule{3-12}
                        &                           & F1           & IoU          & F1            & IoU          & F1          & IoU         & F1           & IoU
                        & F1           & IoU        \\         
\midrule                    
CAT-Net \cite{kwon2022learning}                 & \multirow{3}{*}{FB} & 63.3   & 55.9    & 91.8   &90.0   & 15.1   &11.9   & 12.1   &9.8 &47.4 &42.2     \\
TruFor \cite{guillaro2023trufor}                &                     &67.2    &60.5     & 74.9   &67.1   &35.3    &27.8   & \textcolor{red}{\textbf{65.4}}   &\textcolor{red}{\textbf{55.2}} &57.5 &50.2     \\
% LoMa                 &                     &\textcolor{red}{\textbf{74.7}}    &\textcolor{red}{\textbf{67.4}}     &88.6   &84.9   &\textcolor{red}{\textbf{45.9}}    &\textcolor{red}{\textbf{38.3}}   &\textcolor{red}{\textbf{41.3}}    &\textcolor{red}{\textbf{30.1}}     \\
Ours                                          &                     &\textcolor{red}{\textbf{70.3}}    &\textcolor{red}{\textbf{64.1}}    &\textcolor{red}{\textbf{95.9}}   &\textcolor{red}{\textbf{95.2}}   &\textcolor{red}{\textbf{43.1}}    & \textcolor{red}{\textbf{35.9}}  &39.4  &28.8  &\textcolor{red}{\textbf{62.3}} &\textcolor{red}{\textbf{56.1}}  \\ 
\midrule
CAT-Net \cite{kwon2022learning}                 & \multirow{3}{*}{WC} & 13.9   & 10.6    & 84.8   &80.8   & 19.1   & 14.9  & 1.7   &1.0  &21.4 &17.9   \\
TruFor \cite{guillaro2023trufor}                &                     & 56.9   & 50.8    & 77.3   &70.3   & 35.1   & 27.4  & \textcolor{red}{\textbf{43.6}}   &\textcolor{red}{\textbf{31.4}}          &51.0 &43.9    \\
% LoMa                 &                     &64.2    &56.0     &89.0   &85.3   &45.7    &38.0   &43.0    &31.5     \\
Ours                                            &                     & \textcolor{red}{\textbf{60.9}}   & \textcolor{red}{\textbf{54.1}}    &\textcolor{red}{\textbf{94.6}}   &\textcolor{red}{\textbf{93.5}}  &\textcolor{red}{\textbf{43.5}}      &\textcolor{red}{\textbf{37.0}}  &39.9 &29.4 &\textcolor{red}{\textbf{57.2}} &\textcolor{red}{\textbf{50.8}}    \\ 
\midrule
CAT-Net \cite{kwon2022learning}                 & \multirow{3}{*}{WB} &42.5    &36.2     & 92.1   & 89.7  & 20.8   & 16.0  & 2.3   &1.3 &37.7&32.6      \\
TruFor \cite{guillaro2023trufor}                &                     &63.7    &57.6     & 80.0   & 73.1  & 33.2   & 26.2  & \textcolor{red}{\textbf{46.4}}   &\textcolor{red}{\textbf{36.3}} &54.3 &47.6     \\
% LoMa                 &                     &73.5    &66.6     &88.7   &85.0   &46.4    &39.0   &43.6    &32.0     \\
Ours                                            &                     &\textcolor{red}{\textbf{70.3}}    &\textcolor{red}{\textbf{64.6}}     &\textcolor{red}{\textbf{94.2}}    &\textcolor{red}{\textbf{93.1}}  &\textcolor{red}{\textbf{44.8}}   & \textcolor{red}{\textbf{38.0}}  & 40.6   &29.8  &\textcolor{red}{\textbf{62.5}} &\textcolor{red}{\textbf{56.6}}        \\ 
\midrule
CAT-Net \cite{kwon2022learning}                 & \multirow{3}{*}{WA} & 42.3   & 37.8    & 92.1   &89.9   & 20.1   & 16.8  & 2.2   &1.5  &37.4 &33.7            \\
TruFor \cite{guillaro2023trufor}                &                     & 66.3   & 59.9    & 74.7   &66.7   & 32.3   & 25.6  & 37.6   &28.9 &54.4 &47.7                \\
% LoMa                 &                     &73.7    &66.4     &88.9   &85.2   &47.0    &39.8   &43.0    &31.4     \\
Ours                                            &                     & \textcolor{red}{\textbf{69.7}}   & \textcolor{red}{\textbf{63.6}}    & \textcolor{red}{\textbf{94.1}}   & \textcolor{red}{\textbf{92.9}}  &\textcolor{red}{\textbf{45.1}}    &\textcolor{red}{\textbf{38.9}}  & \textcolor{red}{\textbf{39.8}}   &\textcolor{red}{\textbf{29.6}} &\textcolor{red}{\textbf{62.3}} &\textcolor{red}{\textbf{56.4}}    \\
\bottomrule
\end{tabular}
\end{adjustbox}
\label{osn}
\end{table}

\section{Conclusion}
In this letter, we propose a lightweight image tampering localization method ForMa comprising a visual Mamba encoder, a noise-assisted decoding strategy, and shuffle-based decoder. The Mamba structure efficiently processes inputs while adaptively capturing long-range dependencies, with auxiliary tampering trace features enhancing localization accuracy. The decoder employs pixel shuffle operation to maintain computational efficiency. Experiments demonstrate ForMa's superiority over state-of-the-art CNN- and Transformer-based methods. Future work will focus on improving the model's robustness and generalization capabilities.
\label{sec:conclusion}

\bibliographystyle{IEEEtran}
\bibliography{ref}

\end{document}